\documentclass[11pt, a4paper, logo, copyright, nonumbering]{deepseek}
\usepackage[authoryear, sort&compress, round]{natbib}
\usepackage{dblfloatfix}
\usepackage{ulem}
\usepackage{caption}
\usepackage{subcaption}
\usepackage{makecell}
\usepackage[utf8]{inputenc}
\usepackage{dramatist}
\usepackage{xspace}
\usepackage{pifont}
\usepackage{multirow}
\usepackage{tcolorbox}
\usepackage{tabularx}
\usepackage{xltabular}
\usepackage{longtable}
\usepackage{hyperref}

\usepackage{fontawesome}

\AtBeginDocument{

}
\hypersetup{
    colorlinks=true,
    linkcolor=blue,
    citecolor=blue,
    urlcolor=blue,
    filecolor=blue
}
\usepackage{xcolor}
\usepackage{array}
\usepackage{amsfonts}
\usepackage{amsmath}
\usepackage{amssymb}
\usepackage{lineno}
\usepackage{bm}
\usepackage{booktabs}
\usepackage{ragged2e}

\usepackage[bottom]{footmisc}

\usepackage{CJKutf8}
\usepackage{setspace}

\usepackage{dsfont}

\makeatletter
\def\@BTrule[#1]{%
  \ifx\longtable\undefined
    \let\@BTswitch\@BTnormal
  \else\ifx\hline\LT@hline
    \nobreak
    \let\@BTswitch\@BLTrule
  \else
     \let\@BTswitch\@BTnormal
  \fi\fi
  \global\@thisrulewidth=#1\relax
  \ifnum\@thisruleclass=\tw@\vskip\@aboverulesep\else
  \ifnum\@lastruleclass=\z@\vskip\@aboverulesep\else
  \ifnum\@lastruleclass=\@ne\vskip\doublerulesep\fi\fi\fi
  \@BTswitch}
\makeatother

\addto\extrasenglish{
}

 {\begin{list}{}%
         {\setlength{\leftmargin}{#1}}%
         \item[]%
 }
 {\end{list}}
 
\reportnumber{001} 

\renewcommand{\today}{}

\title{
\centering EdgeFM: Efficient Edge Inference for Vision-Language Models
}

\author[]{
Mengling Deng$^{1\ddagger}$, Yuanpeng Chen$^{1\ddagger \text{\scalebox{0.7}{\faEnvelopeO}} \clubsuit}$, Sheng Yang$^{2\ddagger}$, Wei Tao$^{3}$, Wenhai Zhang$^{1}$,

\vspace{-0.1in}
Hui Song$^{1}$, Linyuanhao Qin$^2$, Kai Zhao$^1$, Xiaojun Ye$^1$, Shanhui Mo$^4$, Jingli Fan$^1$,

\vspace{-0.1in}
Shuang Zhang$^1$, Bei Liu$^1$, Tiankun Zhao$^1$, Xiangjing An$^{1}$ \\
\vspace{-0.1in}
\small
$\ddagger$ Equal Contribution \quad \text{\scalebox{0.7}{\faEnvelopeO}} Corresponding Author \quad $\clubsuit$ Project Leader \\
\small
$^1$Go Further. AI  \quad $^2$School of Data Science, Fudan University\\
\small
$^3$RUYi Dynamics Co., Ltd \quad $^4$Independent Researcher\\
\small
Email: chenyuanpeng@xingshentech.com
\centerline{\small Project Page: \url{https://github.com/windog-labs/edge-fm-x}}
}

\begin{abstract}
Vision-language models (VLMs) have demonstrated strong applicability in edge industrial applications, yet their deployment remains severely constrained by requirements for deterministic low latency and stable execution under resource limitations. Existing frameworks either rely on bloated general-purpose designs or force developers into opaque, hardware-specific closed-source ecosystems, leading to hardware lock-in limitation and poor cross-platform adaptability. Observing that modern AI agents can efficiently search and tune configurations to generate highly optimized low-level kernels for standard LLM operators, we propose \textbf{EdgeFM}, a lightweight, agent-driven VLM/LLM inference framework tailored for cross-platform industrial edge deployment. EdgeFM removes non-essential features to reduce single-request latency, and encapsulates agent-tuned kernel optimizations as a modular library of reusable \textit{skills}. By allowing direct invocation of these skills rather than waiting for closed-source implementations, it effectively closes the performance gap long dominated by proprietary toolchains. The framework natively supports mainstream platforms including x86 and NVIDIA Orin SoCs, and represents \textit{the first end-to-end VLA deployment on the domestic Horizon Journey platform}, enhancing cross-platform portability. In most cases, it yields clearly better inference performance than conventional vendor-specific toolchains, achieving up to $1.49\times$ speedup over TensorRT-Edge-LLM on the NVIDIA Orin platform. Experimental results show that EdgeFM delivers favorable end-to-end inference performance, providing an open-source, production-grade solution for diverse edge industrial scenarios.
\end{abstract}

\begin{document}
\begin{CJK*}{UTF8}{gbsn}

\maketitle

\begin{figure}[htbp]
  \centering
  \includegraphics[width=1\textwidth]{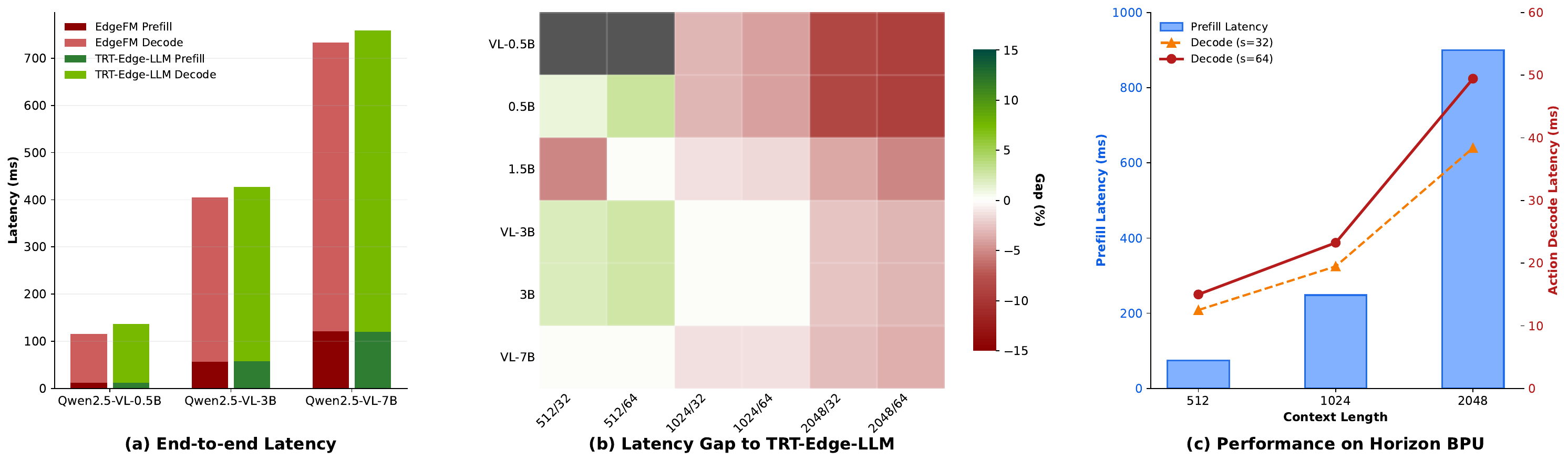}
  
  \caption{\textbf{Overview of cross-platform performance.} (a, b) On the x86 platform: (a) End-to-end latency decomposition; (b) Latency gap under diverse prefill/decode shapes, where red/green indicates EdgeFM/TRT-Edge-LLM advantage; (c) On the Horizon BPU platform: Deployment performance demonstrating deterministic prefill and ultra-low action decode latencies.}
  \label{fig:latency_gap_overall}
\end{figure}


\section{Introduction}
\label{sec:intro}

Large language models~\citep{gpt3, deepseekai2025deepseekr1, qwen2025qwen25technicalreport} (LLMs), built on the attention mechanism~\citep{vaswani2017attention}, have made unprecedented advances in natural language understanding, generation, and reasoning in recent years. However, inference efficiency has become a core bottleneck restricting large-scale deployment across cloud, local, and edge environments. While extensive studies have optimized LLM inference for diverse scenarios, including cloud high-throughput serving and general-purpose local deployment, there remains a critical gap in meeting the stringent requirements of production-grade edge deployment for industrial applications.

Industrial edge scenarios, including autonomous driving and embodied intelligence, impose strict design constraints. Unlike cloud services prioritizing high concurrency, edge workloads are typically single-request and latency-critical, demanding deterministic millisecond-level responsiveness under severely constrained on-board resources. General-purpose local inference frameworks, with their bloated codebases and redundant compatibility layers, struggle to meet these demands. Consequently, industrial edge deployments are frequently forced to rely heavily on proprietary, hardware-vendor-specific closed-source toolchains. While these opaque ecosystems provide baseline performance, they create severe hardware lock-in, act as black boxes that stifle agile algorithm iteration, and force developers into a passive wait for vendor-provided updates whenever novel model architectures emerge.

Fortunately, the remarkable proficiency of modern LLM-based coding agents offers a transformative paradigm shift. Given that the underlying operator semantics of LLM inference are relatively standardized and computationally well-defined, these autonomous agents can assist the optimization workflow of high-performance operators by proposing candidate implementations, tuning kernel parameters, and generating hardware- or shape-specialized artifacts. Crucially, these agent-assisted kernel optimizations are encapsulated into a modular repository of reusable \textit{skills}, while the runtime selects among them through a lightweight operator dispatch table. By directly invoking these skills, developers can attain high-performance inference across diverse hardware platforms without passively waiting for vendor-specific closed-source implementations. This agentic paradigm effectively bridges the performance gap long monopolized by heavily engineered proprietary toolchains, empowering the community to build independent, optimized deployment pipelines free from ecosystem lock-in.

Motivated by this paradigm shift and to eliminate rigid ecosystem dependencies, this paper presents \textbf{EdgeFM} (Edge Foundation Model), a lightweight LLM inference framework natively built for industrial edge platforms. Its core design principles and key contributions are summarized as follows:

\begin{itemize}
    \item \textbf{Edge-native design.} Tailored for single-request edge workloads, EdgeFM enables lightweight deployment via a streamlined API and concise codebase, with a core focus on minimizing end-to-end inference latency.
    \item \textbf{Modular layered architecture.} EdgeFM uses a layer/operator-decoupled design with a model-, hardware-, and stage-aware operator table. Its clean stack of interfaces, lightweight runtime, and unified abstractions dispatches optimal kernels at runtime, ensuring low complexity and independent component evolution.
    \item \textbf{Edge-optimized core acceleration.} EdgeFM integrates mainstream low-latency acceleration techniques. We benchmark against TensorRT-Edge-LLM, NVIDIA’s official and industry-standard framework for high-performance LLM/VLM deployment, and show that our design achieves favorable and consistent latency gains over this strong baseline.
    \item \textbf{Hardware extensibility.} EdgeFM is an open-source, lightweight, lock-in-free framework that supports x86, NVIDIA Orin, and the Horizon Journey 6 series with dedicated adaptations. Moreover, it provides open-source \textit{skills} with reusable agent-tuned kernels for low-cost extension to diverse edge chips and industrial hardware.
\end{itemize}

\section{Related Works}

\subsection{Algorithm-Level Optimizations for LLM Inference}
Algorithm-level improvements form the foundation of efficient LLM inference, focusing on reducing the computational overhead and memory footprint of the autoregressive pipeline. These optimizations apply universally to cloud and edge scenarios. As the core module of LLMs, the attention mechanism can be optimized along two complementary lines: system execution and memory access efficiency, and computational logic and paradigm innovation.

For system execution and cache efficiency optimization, a series of works have broken through the memory wall of attention computation and improved runtime execution efficiency. The Flash Attention series \citep{dao2022flashattention, dao2023flashattention2, shah2024flashattention3, zadouri2026flashattention4} optimizes the memory access pattern of attention computation via kernel fusion and on-chip memory reuse, and has become the de facto standard for high-performance attention implementation. Ring Attention \citep{liu2023ringattention} extends this optimization to distributed scenarios, enabling memory-efficient long-sequence inference across multiple devices. Paged Attention, first proposed in vLLM \citep{kwon2023vllm} and built on the insights of \citet{shazeer2019fast}, introduces OS-inspired paged memory management for KV cache, which greatly improves memory utilization in concurrent inference scenarios.

For computational logic and paradigm optimization, works focus on reducing the inherent computational complexity of the attention mechanism or eliminating redundant computation in practical deployment. Radix Attention, proposed in SGLang \citep{zheng2023sglang}, eliminates redundant prefix computation via radix tree-based KV cache reuse, accelerating the execution of structured language model programs. Sparse attention designs, represented by DeepSeek sparse attention \citep{yuan2025nsa, deepseekai2024deepseekv32}, cut down computational overhead by skipping redundant attention token pairs based on activation sparsity. Linear attention variants \citep{katharopoulos2020transformers}, which are also adopted in industrial systems such as Kimi Linear \citep{team2025kimi}, replace the quadratic complexity of standard attention with linear complexity via kernel transformation, enabling efficient inference for ultra-long sequences.

Weight quantization is another core direction for LLM inference acceleration, reducing memory and computational overhead via low-bit weight compression. As the pioneering works that laid the foundation for LLM weight quantization, LLM.int8() \citep{dettmers2022llmint8} and GPTQ \citep{frantar2023gptq} proposed scalable 8-bit inference and advanced 4-bit post-training quantization for large language models, respectively. As a representative work, AWQ (Activation-aware Weight Quantization) \citep{lin2023awq} achieves high-fidelity compression by aligning quantization with activation distributions, balancing compression ratio, speed and accuracy, and becoming a de facto standard for both cloud and edge systems.

\subsection{System-Level Frameworks for Cloud LLM Serving}
Beyond algorithmic optimizations, cloud and data center serving demands system-level designs to utilize cluster resources for high throughput and concurrency. Industrial-grade inference frameworks further optimize end-to-end performance for dynamic workloads.

vLLM \citep{kwon2023vllm} significantly improves serving throughput through PagedAttention and paged KV cache management \citep{shazeer2019fast}, which efficiently utilize GPU memory under high-concurrency workloads. SGLang \citep{zheng2023sglang} starts from the perspective of structured language model program execution, combining frontend programming abstractions with an efficient runtime to enhance the execution efficiency of complex LM programs. TensorRT-LLM \citep{tensorrtllm} further systematizes and integrates mechanisms such as continuous batching, chunked prefill, KV cache management, and speculative decoding \citep{leviathan2022fast} into a complete industrial-grade inference stack.

At the underlying system level, Orca \citep{yu2022orca} introduces iteration-level scheduling and selective batching, laying a key foundation for fine-grained LLM serving. SARATHI \citep{agrawal2023sarathi} optimizes mixed prefill/decode efficiency via chunked prefills and decode-maximal batching, and FastServe \citep{wu2023fastserve} reduces latency through preemptive scheduling. For decoding acceleration, Speculative Sampling \citep{chen2023speculative} pioneered the draft-verification paradigm to lower autoregressive generation latency, inspiring later studies.

\subsection{Local and Edge-Oriented LLM Inference}

Driven by private, offline and on-device deployment needs, research on general-purpose local LLM inference frameworks has grown rapidly, focusing on model compatibility and cross-platform support. llama.cpp \citep{llamacpp} supports diverse model architectures and hybrid CPU/GPU inference, lowering local deployment barriers. MLC-LLM \citep{mlcllm} enables cross-platform execution via ML compilation, while Candle \citep{candle_hf} provides lightweight, zero-overhead inference for embedded scenarios.

Recent efforts also explore LLM inference in resource-constrained edge environments for edge intelligence. PowerInfer \citep{song2023powerinfer} boosts consumer GPU efficiency via activation locality, and PowerInfer-2 \citep{xue2024powerinfer2} extends it to mobile devices with heterogeneous execution. Tengine LLM \citep{tengine_llm} targets industrial embedded and automotive scenarios with efficient quantization and memory management. MNN-LLM \citep{wang2025mnnllm} optimizes mobile inference via quantization and hybrid storage. These works verify the feasibility of edge LLM deployment and deliver targeted optimizations for diverse edge hardware and workloads.
\section{Methodology}
\label{sec:method}

\subsection{Problem Formulation}

We consider autoregressive generation on an edge device for either a pure language model (LLM) or a vision-language model (VLM). Given an input prompt represented as a token sequence,
\begin{equation}
\mathbf{x}_{1:T} = (x_1, x_2, \ldots, x_T),
\end{equation}
the model generates the output sequence $\mathbf{y}_{1:N}$ token-by-token according to the conditional probability:
\begin{equation}
y_t \sim p_\theta(\cdot \mid \mathbf{x}_{1:T}, \mathbf{y}_{1:t-1}).
\end{equation}

EdgeFM is designed for the typical setting where:
\begin{itemize}
    \item The batch size is typically set to $B=1$;
    \item The system prioritizes tail latency (time-to-first-token, TTFT) and inference predictability;
    \item For edge platforms, device memory is highly constrained, with the KV cache dominating the runtime memory footprint;
    \item Model deployment requires minimal orchestration and operational complexity.
\end{itemize}
Accordingly, EdgeFM optimizes for a deterministic, low-overhead runtime rather than pursuing cloud-style throughput maximization via continuous batching and global scheduling.

\begin{figure}[!t]
  \centering
  \includegraphics[width=\columnwidth]{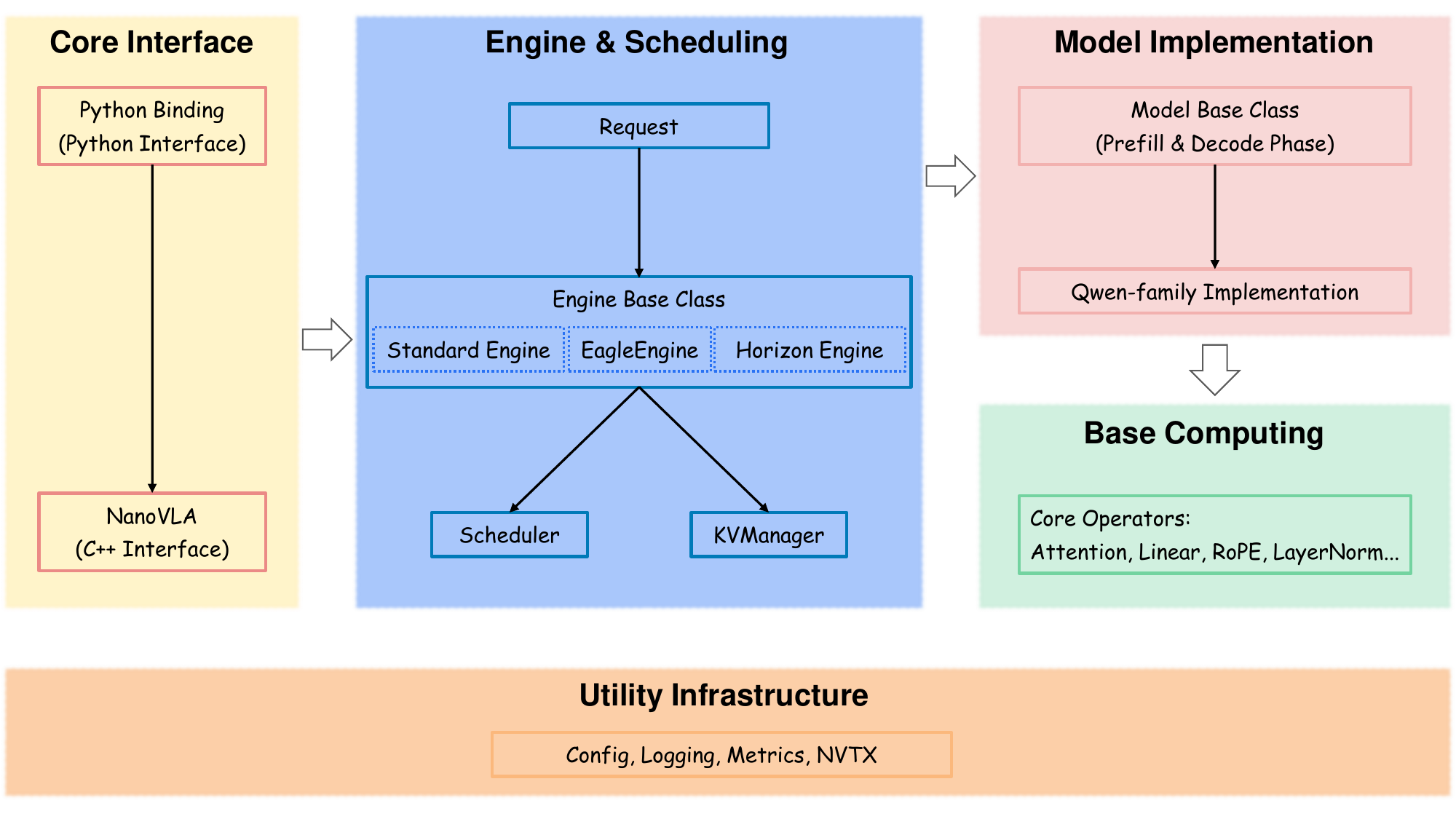}
    \caption{\textbf{Overall architecture of EdgeFM.} In the Engine \& Scheduling Layer, the Scheduler is responsible for request dispatch and single-request batch matching, while the KVManager implements KV cache pre-allocation and compression to improve memory efficiency. The Base Computing Layer further includes three key optimization features: shape-specific operator optimization, advanced high-performance operators, and vocabulary pruning optimization.}
  \label{fig:edgefm_arch}
\end{figure}

\subsection{Design Principles}
\label{subsec:design_principles}

EdgeFM is designed with following core principles for edge-optimized inference:
\begin{itemize}
    \item \textbf{Thin runtime, heavy kernels}: A lightweight core runtime minimizes overhead with optimized hardware-specific kernels, while deterministic single-request execution guarantees stable and predictable latency for edge scenarios.
    \item \textbf{Default implementation first, specialized override}: Universal default implementations enable out-of-the-box usage, and \texttt{operator\_impl\_table} provides targeted specialization only when profitable.
\end{itemize}

\subsection{System Overview}
As shown in Figure~\ref{fig:edgefm_arch}, EdgeFM is a configuration-driven inference engine consisting of the following key elements:
\begin{itemize}
    \item A C++ runtime core responsible for request orchestration, memory management, and model execution.
    \item A CUDA-accelerated operator backend optimized for attention mechanisms and GEMM-heavy compute kernels.
    \item Built-in speculative decoding to further accelerate inference speed.
    \item Lightweight Python bindings for rapid prototyping and system integration.
\end{itemize}
The end-to-end behavior of the framework is defined by a single JSON configuration file, which governs sampling strategies, KV cache policies, and runtime execution schemes.

\noindent \textbf{Request/Response Abstraction.}
A standard inference request is identified by a request ID and accompanied by an input token sequence. The inference engine outputs generated token IDs along with intermediate statistics, including runtime latency measurements and total token counts. This unified design facilitates reproducible evaluation, as each experiment can be defined by a combination of model artifacts, configurations, and input data.

\subsection{Operator Backend Strategy}
EdgeFM follows the \textit{thin runtime, heavy kernels} design principle: its runtime layer only manages request orchestration and memory layout, while performance-critical operators, including attention, GEMM, and KV cache update, are offloaded to highly optimized implementations. On this basis, EdgeFM integrates a high-performance operator backend centered on FlashInfer-class~\citep{ye2025flashinfer} attention kernels, supplemented with carefully selected tuned decode kernels tailored for edge execution. A KV layout abstraction is further incorporated to support MLA-style~\citep{deepseekv2} representations, which enables efficient processing of long multimodal sequences without raising the complexity of the core scheduler.

To instantiate this architecture on new hardware platforms, the mapping from operator calls to specific kernel implementations is dynamically regulated by a two-phase performance optimization workflow, which ultimately populates the operator implementation table. Rather than relying on static heuristic configurations, EdgeFM adopts a systematic tuning pipeline governed by the principle of resolving deterministic dispatch choices through automated static tuning before investing in high-risk manual kernel development.

The optimization workflow proceeds as follows:
\begin{itemize}
    \item \textbf{Static Closed-Loop Tuning:} EdgeFM first establishes a performance baseline by fixing the model architecture, tensor shapes, and profiling runs, recording absolute metrics for prefill and decode stages. The framework then executes \texttt{engine.tune()} to automatically explore the kernel configuration space and generate an operator table overlay. This phase concludes with regression verification, which consumes the generated table and cache to enforce semantic and performance stability through automated pytest and benchmark gates.
    \item \textbf{Profile-Guided Offline Optimization:} For remaining performance bottlenecks, developers utilize profiling tools such as NVIDIA NSYS or NCU to analyze execution traces and rank hotspots by operator and shape. High-priority kernels are isolated and optimized in a standalone repository through an iterative, human-in-the-loop kernel tuning process. Once the targeted speedup is achieved, the accepted kernel is merged back into the main repository through strict regression and performance gates.
\end{itemize}

To ensure optimal resource allocation throughout this workflow, EdgeFM establishes a strict hotspot priority hierarchy. Attention and memory-bound operations, specifically prefill attention, KV cache write paths, and RoPE, are prioritized at the highest level. These are followed by primary projection operators, including QKV, OProj, GateUp, and DownProj, and subsequent decoding modules, such as decode attention and decode linear layers. Normalization, activation, and sampling layers are categorized at a lower priority. Finally, runtime-level overheads, such as runtime launch latency, CUDA graph construction, and host synchronization, are addressed only when the computational kernels are fully stabilized and optimized.

\subsection{Two-Phase Execution}
EdgeFM decomposes autoregressive generation into two distinct phases: prefill and decode.

\noindent \textbf{Prefill Phase.}
Given an input token sequence $\mathbf{x}_{1:T}$, the prefill phase computes the hidden states of the prompt and initializes the KV cache:
\begin{equation}
\mathrm{KV} \leftarrow \texttt{Prefill}(\mathbf{x}_{1:T}; \theta_p),
\end{equation}
where $\theta_p$ denotes the model parameters (or deployment artifact) used for the prefill stage.
The prefill phase typically dominates the time-to-first-token (TTFT), as it processes the full prompt length $T$ in a single forward pass.

\noindent \textbf{Decode Phase.}
The decode phase generates output tokens iteratively, incrementally updating the KV cache with each new token:
\begin{equation}
y_t \leftarrow \texttt{DecodeStep}(y_{t-1}, \mathrm{KV}; \theta_d), \quad \mathrm{KV} \leftarrow \texttt{UpdateKV}(\mathrm{KV}, y_t).
\end{equation}
EdgeFM optionally supports distinct deployment artifacts for prefill ($\theta_p$) and decode ($\theta_d$). When $\theta_d$ is not explicitly provided, the framework automatically falls back to the prefill artifact, allowing both phases to share the same model instance. This design enables stage-specific optimizations when separate artifacts are available:
\begin{itemize}
    \item Optimal deployment precision for prefill and decode stages, maximizing performance and reducing inference latency.
    \item Compatibility with flexible and diverse model architectures (e.g., VLAs), some of them require separated deployment.
\end{itemize}
This optional two-artifact configuration makes the TTFT--throughput trade-off a tunable parameter without imposing additional complexity on default deployments.

Beyond artifact-level separation, EdgeFM further exploits stage-aware and shape-aware operator selection. The operator implementation table can dispatch to different kernel variants according to the current execution phase as well as the input shape signature. This fine-grained control is particularly effective on edge devices where workloads often exhibit fixed or predictable tensor dimensions, allowing the system to consistently select the most efficient kernel for each concrete shape and stage combination.

\noindent \textbf{Graph-Accelerated Decode.}
Within the decode phase, EdgeFM further reduces runtime overhead by using hardware-optimized execution graphs (e.g., CUDA Graphs for NVIDIA GPUs, Horizon BPU Execution Graphs for Horizon edge platforms) to reuse the execution graph across decoding iterations. This design lowers kernel launch overhead and improves responsiveness, which is particularly important for latency-critical edge scenarios.

\subsection{Multi-Task Compatibility}

While standard autoregressive language models remain the primary workloads for edge deployment, a monolithic inference API is insufficient for the diverse paradigms emerging in embodied AI and robotics. Forcing non-standard models into rigid token-generation pipelines leads to sub-optimal scheduling and poor hardware utilization. To address this, EdgeFM introduces a decoupled task contract design, where the task contract dynamically defines the execution boundaries of the engine, request states, stage artifacts, and planners.

\begin{figure}[!b]
  \centering
  \includegraphics[width=1\textwidth]{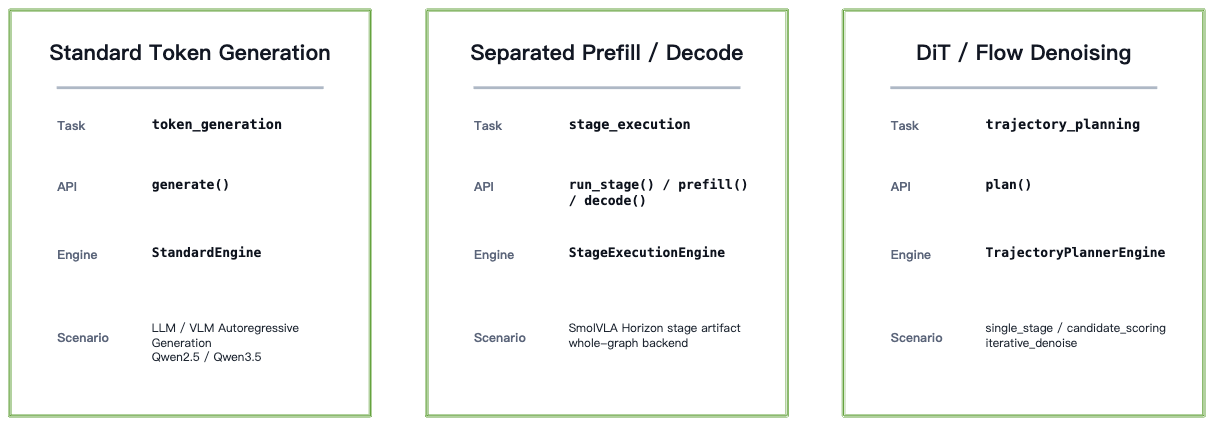}
  \caption{\textbf{Overview of EdgeFM multi-task execution contracts.} Workloads are categorized into three distinct contracts to natively support diverse model architectures without architectural pollution.}
  \label{fig:multitask_contracts}
\end{figure}

As illustrated in Figure~\ref{fig:multitask_contracts}, EdgeFM classifies workloads into three distinct execution contracts, namely \textbf{standard token generation}, \textbf{decoupled stage execution}, and \textbf{trajectory planning}, each managed by a dedicated engine architecture tailored to specific model topologies.

This multi-task separation is driven by two primary motivations. First, while EdgeFM prioritizes highly optimized autoregressive execution as its baseline, it must natively accommodate diverse model families without polluting the core codebase. Second, emerging embodied models such as SmolVLA~\citep{shukor2025smolvla} do not conform to standard, synchronous token generation. SmolVLA relies on a decoupled, stage-wise inference flow where the context prefill phase and the iterative action-denoising loops are highly separated. Forcing these decoupled stages into a synchronous, token-by-token loop would introduce severe runtime overheads and disrupt KV Cache management. Abstracting these behaviors into dedicated engines ensures optimal hardware alignment across all supported model classes.

\subsection{Single-Request Runtime and Scheduling}
\begin{figure}[!t]
  \centering
  \includegraphics[page=1, width=\linewidth]{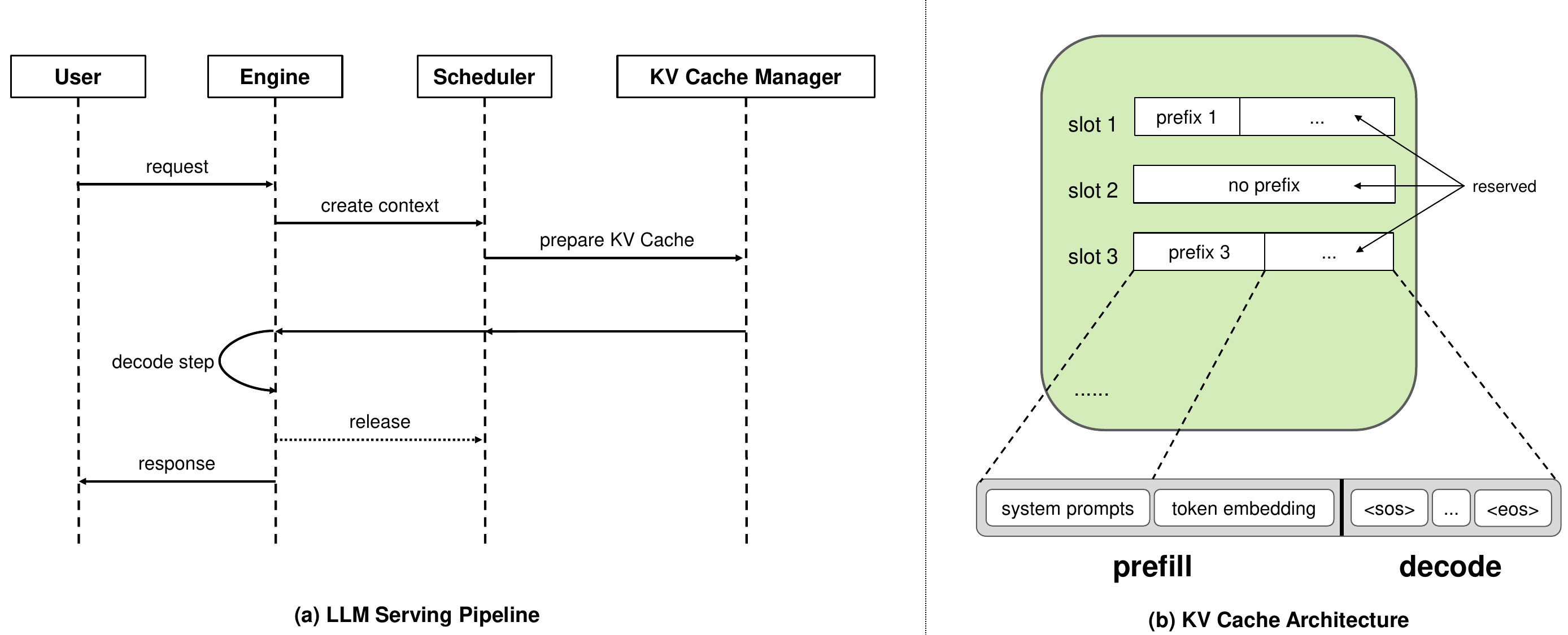}
  \caption{\textbf{EdgeFM LLM serving pipeline and KV cache design.} (a) End-to-end request processing workflow; (b) Slot-based KV cache architecture with prefix reuse as a core optimization in the prefill phase.}
  \label{fig:two_subfigure}
\end{figure}

EdgeFM adopts the single-request execution model. In contrast to continuous batching that aggregates unrelated requests into dynamic micro-batches, EdgeFM processes one request end-to-end with minimal scheduling logic, as illustrated in Figure~\ref{fig:two_subfigure}.

For each request $r$, the runtime executes a fixed, lightweight pipeline:
\begin{itemize}
    \item Allocate or reuse dedicated KV cache slots for the request.
    \item Perform a single prefill pass to process the input prompt.
    \item Run iterative decode steps until stopping criteria are met (EOS, max token limit, etc.).
    \item Release or recycle all occupied resources promptly.
\end{itemize}
This design reduces control-plane overhead and simplifies latency prediction. The methodology aligns with edge deployments where concurrency is limited and predictability is prioritized over maximum aggregate throughput.

\subsection{Operator Implementation Table}
\label{sec:impl_table}

The operator implementation table acts as the central dispatch component of EdgeFM, built to tackle edge hardware fragmentation and avoid the inefficiency of hard-coded kernel bindings. Following the \textit{thin runtime, heavy kernels} design principle, it decouples model execution from fixed kernel implementations, while coordinating with highly optimized operators including FlashInfer-class attention kernels~\citep{ye2025flashinfer}, edge-tuned decode kernels, and MLA-style~\citep{deepseekv2} KV layout support for efficient long multimodal sequence processing. Given the relatively fixed hardware generations and standardized LLM architectures in automotive edge platforms, we pre-tune fixed operator configurations for each target platform to ensure optimal implementation across heterogeneous hardware.

Runtime kernel selection is completed by matching the execution context against predefined configuration entries within the table. A representative set of these key entries is provided in Table~\ref{tab:impl_table_keys}, allowing the system to dynamically select the most suitable optimized kernel for flexible and efficient adaptation across heterogeneous edge platforms.

Kernel dispatch follows a specificity hierarchy: concrete entries (e.g., fixed \texttt{shape\_sig}) override generic defaults. Built-in defaults ensure valid kernels for all supported configurations, while external JSON overrides and a built-in \texttt{auto\_tuning} module enable deployment tuning without runtime modifications. Users can enable \texttt{auto\_tuning} either manually or through engine configurations to optimize operator settings for dedicated hardware. This design realizes low-overhead, maintainable kernel selection, well suited to edge scenarios with predictable execution conditions and limited runtime variability.

\begin{table}[!t]
    \centering
    \caption{\textbf{Key dimensions of the operator implementation table.}}
    \label{tab:impl_table_keys}
    \begin{tabular}{l l p{5cm}}
    \toprule
    \textbf{Dimension} & \textbf{Example Value} & \textbf{Description} \\
    \midrule
    \text{model\_name} & \texttt{qwen2\_5} & Target model family. \\
    \text{hw\_profile} & \texttt{cuda\_sm80} & Hardware capability profile. \\
    \text{op\_kind} & \texttt{linear} & Coarse operator category. \\
    \text{layer\_role} & \texttt{fused\_qkv} & Semantic role of the layer. \\
    \text{op\_name} & \texttt{""} & Fine-grained operator identifier. \\
    \text{stage} & \texttt{decode} & Execution phase. \\
    \text{shape\_sig} & \texttt{m=1|in\_features=1536|...} & Shape signature for shape-specific dispatch. \\
    \text{impl\_id} & \texttt{cublasLt} & Selected kernel implementation identifier. \\
    \text{impl\_params} & \texttt{\{"algo\_index": 0\}} & Additional implementation-specific parameters. \\
    \bottomrule
    \end{tabular}
\end{table}

\subsection{KV Cache Optimization}
\noindent \textbf{Fixed-Prefix Cache Reuse.}
Edge applications feature limited task diversity and fixed inference requests, enabling pre-solidification of common request patterns. These scenarios frequently reuse fixed system prompts, instruction templates, and task descriptors as static prefixes. Unlike cloud systems with runtime prefix discovery, EdgeFM adopts a configuration-driven approach as detailed in Figure~\ref{fig:two_subfigure}: request slots are defined at initialization, each bound to a fixed \texttt{request\_id}, prefix token sequence, and maximum generation length.

The prefix token sequence $\mathbf{p}_{1:K}$ for each slot is precomputed offline. During engine startup, the KV cache is pre-allocated per slot, and a warmup pass computes and stores $\mathrm{KV}_p$ for each configured prefix. For an incoming request matching a slot, only the user-specific suffix $\mathbf{u}$ requires online processing, and the complete KV cache is constructed as:
\begin{equation}
\mathrm{KV} = \texttt{Concat}(\mathrm{KV}_p, \mathrm{KV}_u).
\end{equation}
This slot-based, pre-allocated design enables constant-time cache reuse with simplified memory management and eliminates the overhead of dynamic prefix hashing or runtime cache stitching.

\noindent \textbf{KV Cache Compression.}
The KV cache dominates the memory footprint for long-context inference.
EdgeFM supports pluggable KV cache optimization strategies for memory-constrained edge devices, including high-precision low-bit quantization via SageAttention~\citep{zhang2025sageattention} and structured compression via FlashMLA~\citep{flashmla2025}.
We formalize the compression operation as a transformation:
\begin{equation}
\widetilde{\mathrm{KV}} = C(\mathrm{KV}),
\end{equation}
supplemented by on-demand reconstruction or in-compression-space computing based on kernel support. Compression serves as an independent configurable knob for evaluating memory savings, performance impacts, and generation quality in practical deployments.

\subsection{CUDA Graph Acceleration}
For NVIDIA hardware platforms, EdgeFM implements CUDA Graph acceleration for both the prefill and decode stages, which effectively reduces kernel launch overhead and improves end-to-end inference efficiency.

To support deterministic and efficient replay of CUDA Graphs, we perform the following targeted optimizations:
\begin{itemize}
    \item Underlying operators are adapted by storing dynamic parameters in GPU memory buffers, ensuring stable execution during graph replay.
    \item Dedicated kernels are implemented for KV cache management in the decode stage, constructing a fixed computation graph for each decode step to eliminate parameter reset overhead.
\end{itemize}

\subsection{Speculative Decoding}
To boost decode throughput, EdgeFM integrates EAGLE3-style~\citep{li2025eagle3scalinginferenceacceleration} speculative decoding, where a lightweight draft model $\theta_{\text{draft}}$ proposes candidate tokens that are subsequently verified by the base model $\theta_{\text{base}}$. Following the EAGLE3 design, the draft model takes as input not only the current context but also intermediate-layer features extracted from the base model, improving draft accuracy with minimal additional cost.

Given the current context $c_t = (\mathbf{x}_{1:T}, \mathbf{y}_{1:t-1})$, the procedure alternates between block-wise draft generation and token-wise verification:
\begin{itemize}
    \item \textbf{Draft Proposal:} The draft model generates a candidate token block of length $m$:
    \begin{equation}
    \hat{\mathbf{y}}_{t:t+m-1} \leftarrow \texttt{DraftGenerate}(c_t, \text{base\_features}; \theta_{\text{draft}}).
    \end{equation}
    The KV cache for the draft model is managed alongside that of the base model within a unified \texttt{KVManager}, which allocates dedicated space for draft-specific states without interfering with the base model's cache.
    \item \textbf{Verification:} The base model verifies the proposed block token by token. For each position $i$ from $0$ to $m-1$, the base model computes:
    \begin{equation}
    p_i = p_\theta(\hat{y}_{t+i} \mid c_t, \hat{\mathbf{y}}_{t:t+i-1}; \theta_{\text{base}}),
    \end{equation}
    and accepts $\hat{y}_{t+i}$ only if it matches the base model's greedy prediction or passes a speculative sampling criterion. Verification stops at the first rejection.
    \item \textbf{Commit:} All consecutively accepted tokens are appended to the output sequence, and the base model's KV cache is advanced accordingly. If a rejection occurs, the draft's state is discarded and the next draft block is generated from the updated base context.
\end{itemize}
\textbf{Note:} Speculative pipeline can coexist with both graph-acceleration and fixed-prefix KV reuse.

\section{Experimental Results}

\subsection{Performance Evaluation on x86 Edge Platform}
\label{subsec:x86_performance}

We evaluate the inference performance of our EdgeFM framework on an x86-based edge server equipped with an NVIDIA A800 GPU with 80 GB memory. We benchmark our method against TensorRT-Edge-LLM, the official inference optimization framework developed by NVIDIA, which serves as the industry-standard baseline for high-performance LLM and VLM deployment. The overall experimental results are summarized in Figure~\ref{fig:latency_gap_overall}, and detailed per-shape latency values and their decompositions are provided in Section~\ref{sec:more_tables}. Experiments are conducted across both short and long context scenarios to comprehensively validate the efficiency and robustness of our proposed optimizations.

Across all evaluated configurations, the end-to-end inference latency scales linearly with decode length for both inference frameworks. On the Qwen2.5 model family (including both VLMs and LLMs), EdgeFM consistently achieves lower end-to-end latency than NVIDIA’s official TRT-Edge-LLM under most prefill/decode settings, while maintaining stable and predictable throughput across both short and long context scenarios. These results demonstrate that our design delivers reliable and consistent efficiency gains for real-world LLM/VLM deployment on server-grade edge accelerators.

\subsection{Performance Evaluation on Orin Platform}
\label{subsec:orin_performance}

To further validate the effectiveness of our framework in resource-constrained edge scenarios, we conduct performance evaluation on the NVIDIA Jetson Orin NX 8GB platform. We enforce a strictly limited memory budget for model inference to simulate real-world edge deployment constraints, and benchmark EdgeFM against TensorRT-Edge-LLM on Qwen2.5-VL-0.5B. The overall experimental results are summarized in Table~\ref{tab:orin_performance}. All experiments adopt greedy decoding ($\tau=0$, $\texttt{top\_k}=1$, $\texttt{top\_p}=1$), and each run is validated to generate exactly 32 tokens to ensure measurement accuracy under typical inference scenarios.

Benefiting from the dedicated optimizations for edge hardware, EdgeFM achieves up to $1.49\times$ and $1.38\times$ speedup over TensorRT-Edge-LLM under the 512 and 1024 prefill configurations, respectively. Notably, compared with the results on the x86 server-grade platform, EdgeFM exhibits a more pronounced acceleration advantage on edge devices, verifying the high adaptability of our design to resource-limited hardware.

In particular, this performance advantage is primarily concentrated in the decode stage, where EdgeFM reduces latency by up to $34.35\%$ compared to TensorRT-Edge-LLM. Further analysis shows that prefill latency scales approximately linearly with input length for both frameworks, while decode latency remains nearly constant across different prefill lengths, which is consistent with the characteristics of KV cache reuse. These results demonstrate that our proposed optimizations are particularly effective for decode-bound workloads on resource-limited edge hardware.

\begin{table}[htbp]
\centering
\caption{Performance comparison of EdgeFM and TRT-Edge-LLM on Qwen2.5-VL-0.5B}
\label{tab:orin_performance}
\resizebox{\linewidth}{!}{%
\begin{tabular}{lccccccccc}
\toprule
Shape & \makecell{EdgeFM \\ prefill} & \makecell{TRT \\ prefill} & \makecell{Prefill \\ gap} & \makecell{EdgeFM \\ decode} & \makecell{TRT \\ decode} & \makecell{Decode \\ gap} & \makecell{EdgeFM \\ total} & \makecell{TRT \\ total} & \makecell{Total \\ gap} \\
\midrule
512/32  & 42.5 & 49.5 & -14.14\% & 381.1 & 580.5 & -34.35\% & 423.6 & 630.0 & -32.76\% \\
1024/32 & 86.5 & 90.7 & -4.63\%  & 401.0 & 583.5 & -31.28\% & 487.5 & 674.2 & -27.69\% \\
\bottomrule
\end{tabular}%
}
\end{table}

\subsection{Performance Evaluation on Horizon Edge Platform}
\label{subsec:horizon_performance}

To address the pervasive ecosystem lock-in characteristic of dominant edge inference solutions, we validate the cross-platform adaptability of our framework on the Horizon Journey 6M. This domestic edge SoC is purpose-built for robotics and embodied AI applications where hardware-level optimization is typically constrained by vendor-specific toolchains. On this hardware, we achieve the first end-to-end deployment of a Vision-Language-Action (VLA) model, with the widely adopted SmolVLA-base~\citep{shukor2025smolvla} serving as a representative paradigm. By operating successfully under the platform’s native 5GB BPU memory constraint, this milestone deployment proves that our framework can effectively decouple advanced AI capabilities from specific hardware vendors. Such results demonstrate that EdgeFM provides a viable open-source alternative for high-performance deployment on domestic edge chips.

As evidenced by the results summarized in Figure~\ref{fig:latency_gap_overall}, EdgeFM delivers exceptional inference efficiency. It maintains deterministic low latencies during the computationally intensive prefill stage even as context lengths increase, which ensures that the initial perception phase does not become a bottleneck for downstream execution. More importantly, within the action expert decoding stage that governs the real-time responsiveness of embodied agents, EdgeFM sustains ultra-low latencies for the generation of complete action sequences. These performance metrics collectively confirm that our framework satisfies the stringent requirements for continuous embodied control in complex real-world environments. This high level of efficiency enables the system to support the high-frequency control loops that are essential for maintaining the stability of robotic platforms during dynamic tasks.

\noindent \textbf{Implementation Details on Horizon BPU:} To address current limitations in custom operator support on the BPU, we adopt a decoupled model export strategy for the Horizon platform. We partition the computational execution graph into two distinct static models with explicit management of key-value cache inputs and outputs. Specifically, the prefill model processes the initial prompt and generates the corresponding cache tensors, which the decode model then uses for autoregressive generation. This decoupling ensures each inference phase is fully optimized for the BPU’s tiled execution architecture. We also implement functionally equivalent operator substitutions within the SmolVLA architecture to enable optimal compilation and hardware utilization. Further implementation details are provided in Section~\ref{sec:horizon_details}.
\section{Conclusion}
\label{sec:conclusion}

This work presents EdgeFM, the first open-source lightweight LLM inference framework natively built for cross-platform industrial edge deployment. It breaks free from proprietary hardware ecosystem lock-in while maintaining high deployment efficiency. With edge-centric optimizations prioritizing resource efficiency and scenario-tailored acceleration modules, EdgeFM delivers excellent low-latency inference to meet industrial-grade deployment demands.

Looking ahead, we plan to advance EdgeFM along three directions:

\begin{itemize}
    \item Extend the framework to support end-to-end inference optimization for more VLMs, VLAs, and speech-centric large models (e.g., Qwen3-TTS~\citep{Qwen3-TTS}), enabling efficient multimodal understanding and reasoning for edge scenarios.
    \item Expand hardware support to a wider range of commercial edge platforms, including full-series Horizon Journey 6 chips, with dedicated operator optimizations to further improve efficiency and compatibility.
    \item Integrate dynamic inference scheduling to build a full-stack solution for resource-constrained edge environments and efficient multi-model concurrent deployment.
\end{itemize}

\section*{Statements}
We conduct all experiments using official open-source models from ModelScope and Hugging Face, including

\noindent \textbf{Qwen2.5:} \url{https://modelscope.cn/collections/Qwen25-dbc4d30adb768}

\noindent \textbf{SmolVLA:} \url{https://huggingface.co/lerobot/smolvla_base}

\clearpage
\appendix
\section{Detailed Results}
\label{sec:more_tables}

Here are the detailed latency comparison results (unit: ms) for models in the Qwen2.5 model family on one NVIDIA A800 80GB GPU.

All input shapes are presented in the format of prefill/decode. Besides vision-language models (VLMs), we also report performance benchmarks for large language models (LLMs) from the same series.

\begin{table*}[htbp]
  \centering
  \setlength{\tabcolsep}{3pt}
  \caption{Performance comparison of EdgeFM and TRT-Edge-LLM on Qwen2.5-VL-0.5B}
  \label{tab:qwen25_vl_05b_perf}
  \resizebox{\linewidth}{!}{%
  \begin{tabular}{lcccccccccc}
    \toprule
    Shape &
    \makecell{EdgeFM\\prefill} &
    \makecell{TRT\\prefill} &
    \makecell{Prefill\\gap} &
    \makecell{EdgeFM\\decode} &
    \makecell{TRT\\decode} &
    \makecell{Decode\\gap} &
    \makecell{EdgeFM\\total} &
    \makecell{TRT\\total} &
    \makecell{Total\\gap} \\
    \midrule
    1024/32 & 6.847 & 10.725 & -36.16\% & 53.050 & 55.407 & -4.25\% & 60.042 & 66.240 & -9.36\% \\
    1024/64 & 6.289 & 8.032 & -21.70\% & 104.542 & 112.548 & -7.11\% & 111.093 & 120.713 & -7.97\% \\
    2048/32 & 12.314 & 12.188 & +1.03\% & 50.626 & 60.965 & -16.96\% & 63.026 & 73.228 & -13.93\% \\
    2048/64 & 12.299 & 12.235 & +0.52\% & 103.542 & 123.893 & -16.43\% & 115.968 & 136.226 & -14.87\% \\
    \bottomrule
  \end{tabular}%
  }
\end{table*}

\begin{table*}[htbp]
  \centering
  \caption{Performance comparison of EdgeFM and TRT-Edge-LLM on Qwen2.5-VL-3B}
  \label{tab:qwen25_vl_3b_perf}
    \resizebox{\linewidth}{!}{%
  \begin{tabular}{lccccccccccc}
    \toprule
    Shape &
    \makecell{EdgeFM\\prefill} &
    \makecell{TRT\\prefill} &
    \makecell{Prefill\\gap} &
    \makecell{EdgeFM\\decode} &
    \makecell{TRT\\decode} &
    \makecell{Decode\\gap} &
    \makecell{EdgeFM\\total} &
    \makecell{TRT\\total} &
    \makecell{Total\\gap} \\
    \midrule
    512/32  & 16.055 & 17.798 & -9.79\% & 169.883 & 164.621 & +3.20\% & 185.938 & 182.419 & +1.93\% \\
    512/64  & 16.050 & 17.739 & -9.52\% & 345.357 & 334.481 & +3.25\% & 361.406 & 352.220 & +2.61\% \\
    1024/32 & 28.661 & 29.281 & -2.12\% & 170.918 & 170.133 & +0.46\% & 199.579 & 199.414 & +0.08\% \\
    1024/64 & 28.659 & 29.191 & -1.82\% & 346.527 & 345.751 & +0.22\% & 375.186 & 374.942 & +0.07\% \\
    2048/32 & 56.011 & 57.682 & -2.90\% & 172.185 & 182.960 & -5.89\% & 228.196 & 240.642 & -5.17\% \\
    2048/64 & 56.222 & 57.645 & -2.47\% & 349.029 & 369.865 & -5.63\% & 405.250 & 427.510 & -5.21\% \\
    \bottomrule
  \end{tabular}
  }
\end{table*}

\begin{table*}[htbp]
  \centering
  \caption{Performance comparison of EdgeFM and TRT-Edge-LLM on Qwen2.5-VL-7B}
  \label{tab:qwen25_vl_7b_perf}
    \resizebox{\linewidth}{!}{%
  \begin{tabular}{lccccccccccc}
    \toprule
    Shape &
    \makecell{EdgeFM\\prefill} &
    \makecell{TRT\\prefill} &
    \makecell{Prefill\\gap} &
    \makecell{EdgeFM\\decode} &
    \makecell{TRT\\decode} &
    \makecell{Decode\\gap} &
    \makecell{EdgeFM\\total} &
    \makecell{TRT\\total} &
    \makecell{Total\\gap} \\
    \midrule
    512/32  & 30.525 & 29.883 & +2.15\% & 300.683 & 300.502 & +0.06\% & 331.207 & 330.384 & +0.25\% \\
    512/64  & 30.572 & 29.798 & +2.60\% & 610.981 & 610.172 & +0.13\% & 641.553 & 639.970 & +0.25\% \\
    1024/32 & 58.343 & 57.854 & +0.85\% & 300.932 & 305.951 & -1.64\% & 359.276 & 363.805 & -1.25\% \\
    1024/64 & 58.648 & 58.522 & +0.22\% & 611.449 & 619.784 & -1.34\% & 670.097 & 678.307 & -1.21\% \\
    2048/32 & 120.855 & 120.110 & +0.62\% & 301.360 & 315.002 & -4.33\% & 422.215 & 435.112 & -2.96\% \\
    2048/64 & 120.841 & 120.290 & +0.46\% & 612.216 & 638.760 & -4.16\% & 733.056 & 759.050 & -3.42\% \\
    \bottomrule
  \end{tabular}
  }
\end{table*}

\begin{table*}[htbp]
  \centering
  \caption{Performance comparison of EdgeFM and TRT-Edge-LLM on Qwen2.5-0.5B}
  \label{tab:qwen25_05b_perf}
    \resizebox{\linewidth}{!}{%
  \begin{tabular}{lccccccccccc}
    \toprule
    Shape &
    \makecell{EdgeFM\\prefill} &
    \makecell{TRT\\prefill} &
    \makecell{Prefill\\gap} &
    \makecell{EdgeFM\\decode} &
    \makecell{TRT\\decode} &
    \makecell{Decode\\gap} &
    \makecell{EdgeFM\\total} &
    \makecell{TRT\\total} &
    \makecell{Total\\gap} \\
    \midrule
    512/32  & 3.671 & 8.061 & -54.46\% & 56.547 & 51.452 & +9.90\% & 60.217 & 59.513 & +1.18\% \\
    512/64  & 3.509 & 7.038 & -50.14\% & 111.479 & 104.882 & +6.29\% & 114.988 & 111.920 & +2.74\% \\
    1024/32 & 6.374 & 8.501 & -25.02\% & 54.924 & 54.720 & +0.37\% & 61.298 & 63.221 & -3.04\% \\
    1024/64 & 6.088 & 9.281 & -34.41\% & 109.560 & 111.206 & -1.48\% & 115.648 & 120.487 & -4.02\% \\
    2048/32 & 11.886 & 11.820 & +0.56\% & 54.134 & 60.262 & -10.17\% & 66.020 & 72.082 & -8.41\% \\
    2048/64 & 11.866 & 11.930 & -0.53\% & 110.004 & 122.215 & -9.99\% & 121.870 & 134.144 & -9.15\% \\
    \bottomrule
  \end{tabular}
  }
\end{table*}

\begin{table*}[htbp]
  \centering
  \caption{Performance comparison of EdgeFM and TRT-Edge-LLM on Qwen2.5-1.5B}
  \label{tab:qwen25_15b_perf}
    \resizebox{\linewidth}{!}{%
  \begin{tabular}{lccccccccccc}
    \toprule
    Shape &
    \makecell{EdgeFM\\prefill} &
    \makecell{TRT\\prefill} &
    \makecell{Prefill\\gap} &
    \makecell{EdgeFM\\decode} &
    \makecell{TRT\\decode} &
    \makecell{Decode\\gap} &
    \makecell{EdgeFM\\total} &
    \makecell{TRT\\total} &
    \makecell{Total\\gap} \\
    \midrule
    512/32  & 8.607 & 16.142 & -46.68\% & 100.137 & 98.747 & +1.41\% & 108.744 & 114.888 & -5.35\% \\
    512/64  & 8.613 & 11.076 & -22.24\% & 203.570 & 200.499 & +1.53\% & 212.182 & 211.575 & +0.29\% \\
    1024/32 & 15.294 & 15.710 & -2.65\% & 101.320 & 102.646 & -1.29\% & 116.613 & 118.356 & -1.47\% \\
    1024/64 & 15.303 & 16.252 & -5.84\% & 205.955 & 208.512 & -1.23\% & 221.259 & 224.764 & -1.56\% \\
    2048/32 & 30.073 & 29.405 & +2.27\% & 104.723 & 110.739 & -5.43\% & 134.796 & 140.144 & -3.82\% \\
    2048/64 & 29.993 & 29.990 & +0.01\% & 211.888 & 225.484 & -6.03\% & 241.881 & 255.474 & -5.32\% \\
    \bottomrule
  \end{tabular}
  }
\end{table*}

\begin{table*}[htbp]
  \centering
  \caption{Performance comparison of EdgeFM and TRT-Edge-LLM on Qwen2.5-3B}
  \label{tab:qwen25_3b_perf}
    \resizebox{\linewidth}{!}{%
  \begin{tabular}{lccccccccccc}
    \toprule
    Shape &
    \makecell{EdgeFM\\prefill} &
    \makecell{TRT\\prefill} &
    \makecell{Prefill\\gap} &
    \makecell{EdgeFM\\decode} &
    \makecell{TRT\\decode} &
    \makecell{Decode\\gap} &
    \makecell{EdgeFM\\total} &
    \makecell{TRT\\total} &
    \makecell{Total\\gap} \\
    \midrule
    512/32  & 16.057 & 17.386 & -7.64\% & 171.640 & 166.648 & +3.00\% & 187.697 & 184.034 & +1.99\% \\
    512/64  & 16.087 & 18.218 & -11.70\% & 348.960 & 339.887 & +2.67\% & 365.047 & 358.106 & +1.94\% \\
    1024/32 & 28.767 & 29.943 & -3.93\% & 171.036 & 172.533 & -0.87\% & 199.803 & 202.475 & -1.32\% \\
    1024/64 & 28.760 & 29.280 & -1.78\% & 346.214 & 350.497 & -1.22\% & 374.974 & 379.777 & -1.26\% \\
    2048/32 & 58.110 & 57.025 & +1.90\% & 177.585 & 184.774 & -3.89\% & 235.695 & 241.800 & -2.52\% \\
    2048/64 & 57.872 & 57.220 & +1.14\% & 359.416 & 374.117 & -3.93\% & 417.289 & 431.338 & -3.26\% \\
    \bottomrule
  \end{tabular}
  }
\end{table*}
\clearpage
\section{Implementation Details on Horizon BPU}
\label{sec:horizon_details}

\subsection{Detailed Latency Benchmarks on Horizon BPU}
The detailed latency metrics for SmolVLA-0.45B on the Horizon Journey 6M platform are summarized in Table~\ref{tab:prefill_latency} for the prefill stage and Table~\ref{tab:decode_latency} for the action expert decoding stage.

\begin{table}[htbp]
\centering
\caption{Prefill Stage Latency of SmolVLA-0.45B}
\label{tab:prefill_latency}
\begin{tabular}{cc}
\toprule
\textbf{Prefix} & \textbf{Mean Latency (ms)} \\
\midrule
512 & 74.26 \\
1024 & 248.27 \\
2048 & 899.39 \\
\bottomrule
\end{tabular}
\end{table}

\begin{table}[htbp]
\centering
\caption{Action Expert Decoding Stage Latency}
\label{tab:decode_latency}
\begin{tabular}{cccccc}
\toprule
\textbf{Prefix} & \textbf{Suffix} & \textbf{Mean (ms)} & \textbf{Median (ms)} & \textbf{Min (ms)} & \textbf{Max (ms)} \\
\midrule
512 & 32 & 12.51 & 12.42 & 12.31 & 12.89 \\
512 & 64 & 15.00 & 14.89 & 14.79 & 15.47 \\
1024 & 32 & 19.48 & 19.32 & 19.13 & 20.62 \\
1024 & 64 & 23.25 & 23.13 & 22.91 & 23.62 \\
2048 & 32 & 38.38 & 38.41 & 37.86 & 38.75 \\
2048 & 64 & 49.43 & 49.43 & 48.92 & 50.49 \\
\bottomrule
\end{tabular}
\end{table}

\subsection{Stage Partitioning and KV Cache Management}
Due to the static graph requirements and custom operator constraints of the Horizon BPU, we decouple the SmolVLA inference process into two primary stages: \textit{Prefill} and \textit{Decode}. 
\begin{itemize}
    \item \textbf{Prefill Stage:} The \texttt{smolvla\_prefill.hbm} handles the multimodal input embeddings (images, language, and state). It outputs the Key-Value (KV) cache tensors for all $L$ layers.
    \item \textbf{Decode Stage:} The \texttt{smolvla\_decode.hbm} manages the action expert and Flow Matching denoising steps. It takes the action latent $x_t$, timestep $t$, and the explicitly passed prefix KV cache as inputs, outputting the predicted velocity field $v_t$.
\end{itemize}
This explicit management allows EdgeFM to cache the prefill results in the BPU-accessible memory, avoiding redundant computations during the iterative denoising loop.

\subsection{Operator Substitutions for Numerical Stability}
To ensure optimal compilation and prevent numerical overflow during INT16 quantization, several operators in the original SmolVLA implementation are substituted with hardware-friendly equivalents:

\begin{enumerate}
    \item \textbf{GELU Activation:} The original \texttt{gelu\_pytorch\_tanh} implementation involves an $x^3$ term, which can easily exceed the range of Float16/INT16 during intermediate calculations (e.g., $x=300 \implies x^3 > 2.7 \times 10^7$). We substitute this with the \textit{erf}-based GELU:
    \begin{equation}
        \text{GELU}_{\text{erf}}(x) = 0.5x \left(1 + \text{erf}\left(\frac{x}{\sqrt{2}}\right)\right)
    \end{equation}
    
    \item \textbf{Rotary Positional Embeddings (RoPE):} To adapt to the BPU's compute patterns, we decompose the RoPE logic into a sequence of explicit \texttt{slice}, \texttt{negate}, and \texttt{concat} operations. This ensures that the Horizon compiler can effectively fuse these operations into high-performance kernels.
    
    \item \textbf{Attention Masking:} Standard implementations often use $-\infty$ or \texttt{finfo.min} for masking, which corrupts the quantization scale. We replace these with a safe negative constant $V_{\text{safe}}$:
    \begin{equation}
        \text{Score}_{\text{masked}} = \text{where}(\text{mask}, \text{Score}, V_{\text{safe}})
    \end{equation}
    where $V_{\text{safe}}$ is calibrated to ensure that the masked positions result in zero probability after the Softmax operation without dominating the tensor's dynamic range.
\end{enumerate}

\subsection{INT16 Quantization and Loop-wise Calibration}
We employ a uniform INT16 quantization strategy for both weights and activations to maintain model accuracy. A critical challenge in deploying Flow Matching models like SmolVLA is the shift in activation distribution across different denoising steps. 

To address this, our calibration process does not rely solely on the initial noise ($t=0$). Instead, we collect activation statistics across the entire denoising trajectory (steps $0$ to $N$). This \textit{loop-wise calibration} ensures that the quantization parameters (scales and offsets) are robust across the shifting distributions of the denoising process, preventing error accumulation that could otherwise lead to trajectory divergence.

\subsection{Inference Pipeline Pseudocode}
The EdgeFM framework orchestrates the SmolVLA inference as follows:
\begin{enumerate}
    \item \textbf{Initialization:} Load the decoupled HBM artifacts and initialize the Horizon runtime backend.
    \item \textbf{Contextual Encoding:} Invoke \texttt{engine.prefill(request\_id, inputs)} to generate and cache the prefix KV tensors.
    \item \textbf{Iterative Denoising:} Within the Flow Matching loop, invoke the decoder engine. The backend automatically injects the cached prefix KV tensors into the decode model.
    \item \textbf{Action Generation:} After $N$ iterations, the final latent $x_0$ is de-normalized to produce the robot execution commands.
\end{enumerate}

\clearpage
\bibliography{ref}

\clearpage

\end{CJK*}
\end{document}